\documentclass{article}
\usepackage{spconf,amsmath,graphicx,hyperref}
\usepackage{algorithm}
\usepackage{algpseudocode}
\usepackage{wrapfig}    
\usepackage{graphicx}   
\usepackage[most]{tcolorbox} 
\usepackage{caption}            
\usepackage{lipsum}     
\usepackage{booktabs}
\usepackage{multirow}
\usepackage{spconf,amsmath,amssymb,graphicx,hyperref}
\usepackage{listings}
\usepackage{xcolor}

\definecolor{usergray}{gray}{0.35}
\definecolor{modelblue}{RGB}{0,76,153}
\definecolor{gtgreen}{RGB}{0,120,80}
\definecolor{actionred}{RGB}{150,0,0}

\newcommand{\user}[1]{\textcolor{usergray}{\textbf{User:} #1}}
\newcommand{\model}[1]{\textcolor{modelblue}{\textbf{E-SocialNav:} #1}}
\newcommand{\gt}[1]{\textcolor{gtgreen}{\textbf{Ground truth:} #1}}
\newcommand{\action}[1]{\textcolor{actionred}{#1}}

\title{E-SocialNav: Efficient Socially Compliant Navigation with Language Models}
\name{
  Ling Xiao$^{1}$\thanks{Thanks to JSPS KAKENHI (Grant Number 24K20787) for funding.},
  Daeun Song$^{2}$, Xuesu Xiao$^{2}$,
  Toshihiko Yamasaki$^{3}$
}
\address{
  $^{1}$Hokkaido University, Sapporo, Japan \\
  $^{2}$George Mason University, Virginia, United States \\
  $^{3}$The University of Tokyo, Tokyo, Japan
}

\begin{document}
%
\maketitle
\begin{abstract}
Language models (LMs) are increasingly applied to robotic navigation; however, existing benchmarks primarily emphasize navigation success rates while paying limited attention to social compliance. Moreover, relying on large-scale LMs can raise efficiency concerns, as their heavy computational overhead leads to slower response times and higher energy consumption, making them impractical for real-time deployment on resource-constrained robotic platforms. In this work, we evaluate the social compliance of GPT-4o and Claude in robotic navigation and propose E-SocialNav, an efficient LM designed for socially compliant navigation. Despite being trained on a relatively small dataset, E-SocialNav consistently outperforms zero-shot baselines in generating socially compliant behaviors. By employing a two-stage training pipeline consisting of supervised fine-tuning followed by direct preference optimization, E-SocialNav achieves strong performance in both text-level semantic similarity to human annotations and action accuracy. The source code is available at \url{https://github.com/Dr-LingXiao/ESocialNav}.
\end{abstract}
\section{Introduction}
\label{sec:intro}

Mobile robots fulfill a wide range of functions, from assisting in healthcare and eldercare to providing delivery and logistics services, and supporting security and surveillance tasks. These roles often require robots to interact effectively with humans and to navigate seamlessly through public spaces shared with pedestrians. In such dynamic environments, it becomes crucial for robots to demonstrate socially compliant behaviors in both interaction and navigation, ensuring safety, efficiency, and user acceptance~\cite{payandeh2024social}.

The primary challenges of this task lie in understanding and predicting human intentions, managing uncertainty in dynamic and cluttered environments, and balancing efficiency with safety and comfort. To achieve this, robots need to integrate perception, prediction, and planning modules capable of producing socially compliant trajectories that can adapt to diverse interaction scenarios.

Existing methods include imitation learning (IL)-based~\cite{cuan2024gesture2path}, reinforcement learning (RL)-based~\cite{kathuria2025learning}, and large language model (LLM)-based approaches~\cite{xiao2025llm,song2024vlm}. Among these, LLM-based methods are particularly promising because LLMs provide strong contextual understanding and commonsense reasoning, which align well with the requirements of socially compliant navigation.

Despite recent progress, relying on LLMs may introduce efficiency challenges. For example, VLM-Social-Nav~\cite{song2024vlm} employs GPT-4v to generate navigation instructions; however, due to its large parameter size and the inability to leverage GPU acceleration, this results in significant inference latency. In addition, there has been no systematic evaluation of the zero-shot capabilities of existing LLMs (such as GPT-4 and Claude) for socially aware navigation. Understanding how well off-the-shelf models perform without task-specific training is essential for assessing their readiness for real-world deployment. Building on these insights, it is also critical to design a trainable model that is GPU-accelerated and efficient. Nevertheless, fine-tuning LLMs for this task faces a practical obstacle: high-quality, large-scale datasets are scarce, making it imperative to explore how limited data can be leveraged effectively.

This paper addresses the above-mentioned issues. First, we conduct a comprehensive zero-shot evaluation of GPT-4o and Claude for socially compliant navigation. Second, we propose E-SocialNav, an efficient LM designed for socially compliant navigation under small-data settings. The main contributions are summarized as follows:
\begin{itemize}
    \item We evaluate GPT-4o and Claude, and develop E-SocialNav for efficient navigation under small-data settings.  
    \item We build a multi-dialog SFT dataset and a single-dialog DPO dataset for socially compliant navigation.  
    \item We identify suitable Small Language Models (SLMs) and Vision Towers (VTs) for this task.  
\end{itemize}

\section{Related Work}
\subsection{Social Robot Navigation}
For social robot navigation, safety is paramount~\cite{liang2021crowd}. Classical methods enforce collision constraints or fuse multi-sensor data (2-D LiDAR, depth cameras) for smooth avoidance~\cite{liang2021crowd}.

Safety alone, however, is insufficient in human-populated spaces. Robots must also respect social norms (such as personal space, group dynamics, and cultural conventions) to be perceived as acceptable and trustworthy. Traditional methods often ignore these, reducing pedestrians to moving obstacles.

Learning-based approaches seek to encode both safety and social awareness. Demonstration-driven motion learning~\cite{sun2021motion} and RL in simulators~\cite{liang2021crowd} show promise but depend on large datasets or highly realistic human simulations, which rarely capture nuanced interactions, yielding policies with poor generalization.

Recently, Multimodal Large Language Models (MLLMs) have opened new directions. MLLMs generate high-level actions~\cite{payandeh2024social}, evaluate trajectories~\cite{narasimhan2024olivia}, and predict directions~\cite{song2024vlm}. Datasets such as SCAND~\cite{karnan2022socially} and MuSoHu~\cite{nguyen2023toward} further enable socially compliant, human-like navigation. However, research on small language models (SLMs) for this task remains limited.

\subsection{Small Language Models}

\begin{figure}[t]
\centering
\includegraphics[width=0.5\textwidth]{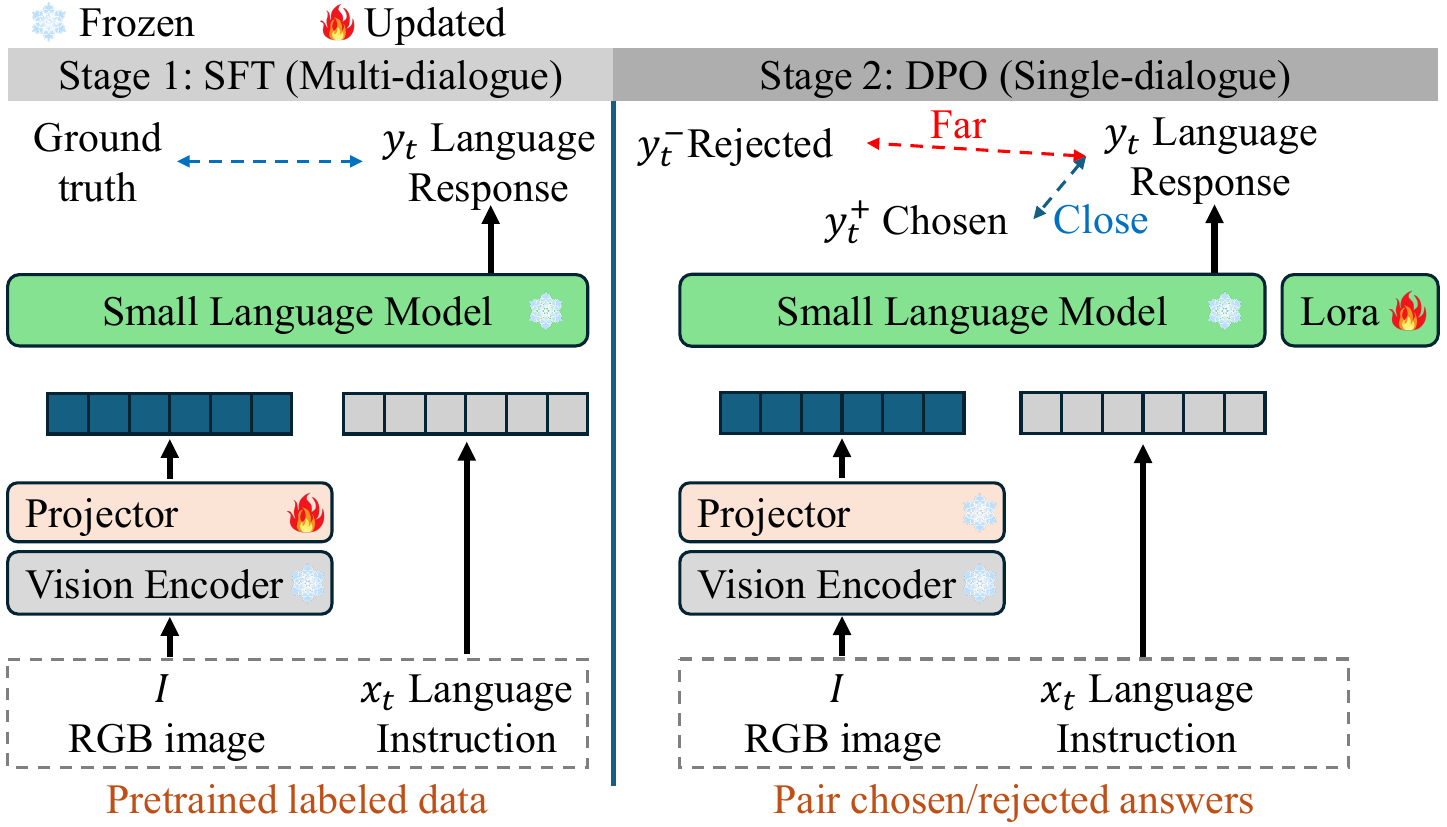}
\caption{The detailed structure of E-SocialNav. E-SocialNav is trained in two phases: SFT on multi-turn dialogues, followed by DPO on single-turn pairs. During SFT, only the projector is updated; during DPO, only the LoRA adapter is updated.}
\label{fig:model}
\end{figure}

LLMs have shown strong abilities in reasoning, planning, and multimodal understanding. While frontier models (e.g., GPT-4, Claude) achieve state-of-the-art performance, their substantial computational demands hinder deployment in robotics and edge devices due to higher inference latency and greater computational consumption.

Recent work therefore emphasizes small language models (SLMs)~\cite{zhou2024tinyllava}. Three main directions have emerged: (1) Efficient pretraining and distillation: transferring knowledge from large teachers via distillation or pruning~\cite{kandala2024tinyllm} to retain reasoning capacity at lower cost; (2) Parameter-efficient fine-tuning: methods such as LoRA~\cite{hulora} and prompt-tuning enable task specialization with minimal overhead; (3) Architectural and training innovations: lightweight models (e.g., TinyLLaMA~\cite{zhang2024tinyllama}) and data-efficient recipes build compact yet capable SLMs. This reflects a shift from pure scaling toward deployability. By aligning efficiency with contextual reasoning, SLMs offer a practical path to bring language models into real-world interactive systems where resources, cost, and latency are critical.

\section{Methods}
\label{sec:method}

\begin{figure}[t]
\centering
\begin{minipage}{.49\linewidth}
  \centering
  \includegraphics[width=\linewidth]{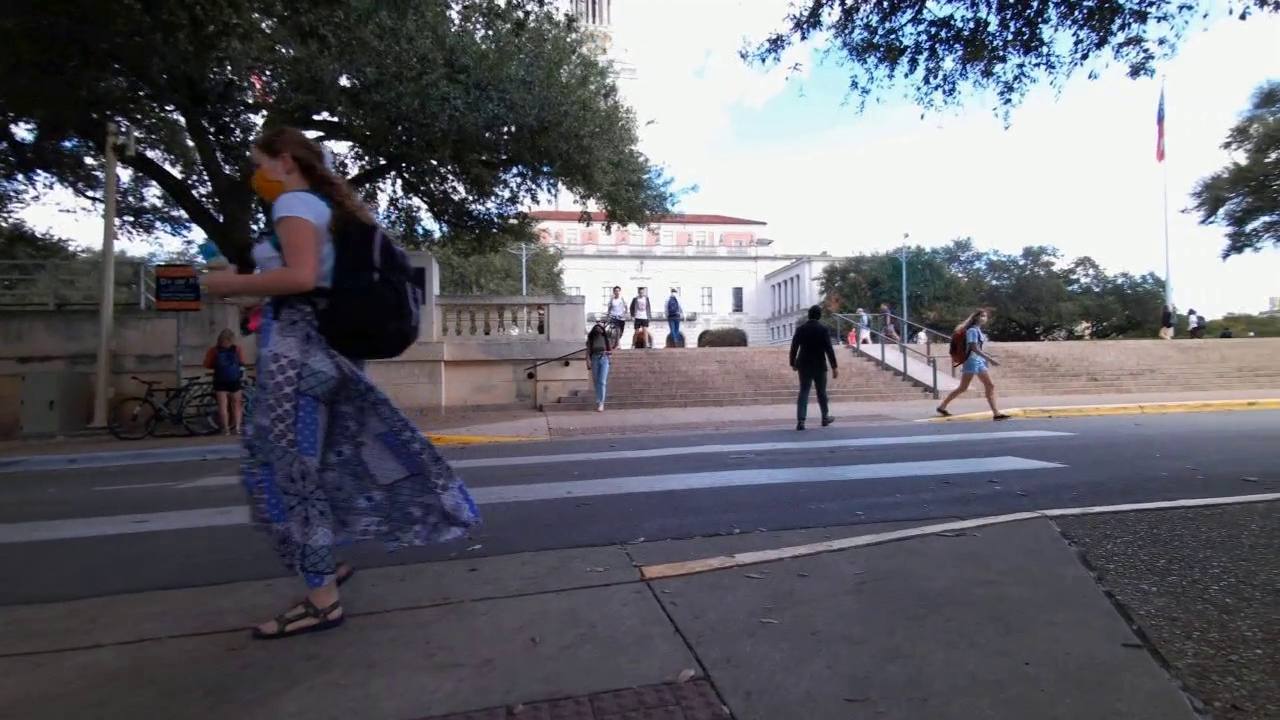}
  \vspace{2pt}
  \scriptsize
  \par\textbf{Human:} ``\texttt{<image>} What should the robot do?'' 
  \par\textbf{Chosen:} ``The robot should stop, wait for clear path.'' 
  \par\textbf{Rejected:} ``The robot should continue straight.'' 
\end{minipage}\hfill
\begin{minipage}{.49\linewidth}
  \centering
  \includegraphics[width=\linewidth]{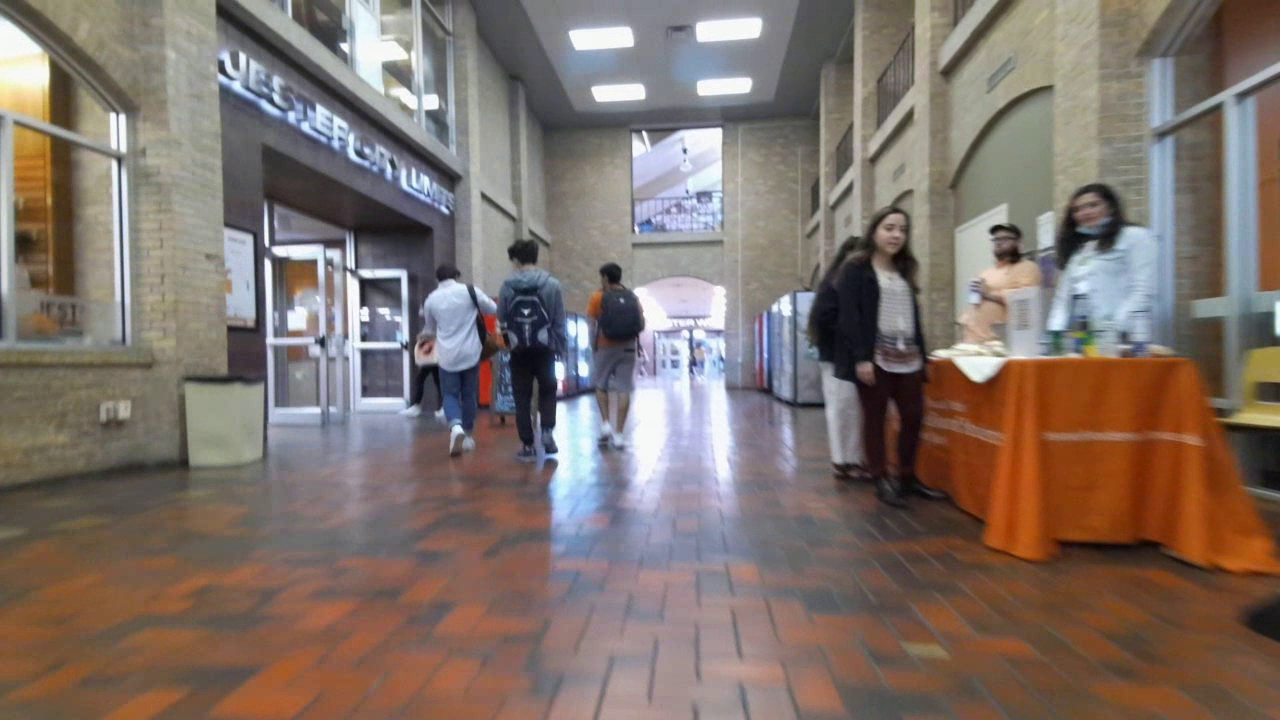}
  \vspace{2pt}
  \scriptsize
  \par\textbf{Human:} ``\texttt{<image>} What should the robot do?'' 
  \par\textbf{Chosen:} ``The robot should continue straight at a moderate speed.'' 
  \par\textbf{Rejected:} ``The robot should stop and wait.'' 
\end{minipage}
\caption{Visualization of constructed DPO training pairs. The chosen response is annotated by humans, whereas the rejected response is generated by modifying certain facts in the chosen response.}
\label{fig:dpo-visual}
\end{figure}

\begin{table*}[t]
\centering
\caption{Experimental results comparing off-the-shelf models and variants of the proposed method. SFT(X) means the components X are trainable in Stage I (supervised fine-tuning); DPO(Y) means Y are trainable in Stage II (direct preference optimization). Components not listed are frozen. Best performance is bolded.}
\label{tab:model_comparison}
\resizebox{\textwidth}{!}{
\begin{tabular}{lcccccccccc}
\toprule
 & Model  &VT &  LM& BERTScore-P$\uparrow$  & BERTScore-R$\uparrow$  & BERTScore-F1$\uparrow$ & SBERT-cos$\uparrow$ & SMS$\uparrow$ & FPS$\uparrow$ & AA$\uparrow$\\
\midrule
\multirow{2}{*}{Off-the-shelf} & Claude  &-& - & -0.233   &  0.387 & 0.059  & 0.664 & 0.641 & 0.087 & 0.417\\
& GPT-4o&-& -   &0.076   & 0.443   & 0.254 & 0.672   &0.651 & 0.212 &  0.450\\  \midrule
Finetuned & Social-LLaVA~\cite{payandeh2024social}  &CLIP ViT-L/14& Vicuna-7B & 0.672 &  0.653  &0.641    &0.784  & 0.813 &  1.113 &  0.483 \\ \midrule
\multirow{4}{*}{E-SocialNav (Ours)} & SFT(projector+lora+vision) & SigLIP~\cite{zhai2023sigmoid} & Phi-2-2.7B~\cite{javaheripi2023phi}  &  0.585  &   0.434  & 0.509    &  0.744 &  0.802  & 0.978  & 0.383\\ 
& SFT(projector+lora) & SigLIP~\cite{zhai2023sigmoid} & Phi-2-2.7B~\cite{javaheripi2023phi}  &   0.640  & 0.545    &    0.592 &  0.756  &   0.813  & 1.553 &  0.400 \\  \cline{2-11}
& SFT(projector) & SigLIP~\cite{zhai2023sigmoid} & Phi-2-2.7B~\cite{javaheripi2023phi}  & 0.551 &  0.658  &0.604    &0.780  & 0.828 &  1.828 &  0.433\\
& SFT(projector) + DPO(lora) & SigLIP~\cite{zhai2023sigmoid} & Phi-2-2.7B~\cite{javaheripi2023phi}  & \textbf{0.706}   &  \textbf{0.671}   &    \textbf{0.688} &   \textbf{0.814}  &  \textbf{0.846}  & \textbf{2.354}  & \textbf{0.550} \\ 
\bottomrule
\end{tabular}
}
\end{table*}

\begin{figure*}[t]
    \centering 
    \begin{tcolorbox}[colback=gray!10!white, colframe=gray!10!black]
    \begin{wrapfigure}{r}{0.2\textwidth}
        \centering
        \includegraphics[width=\linewidth]{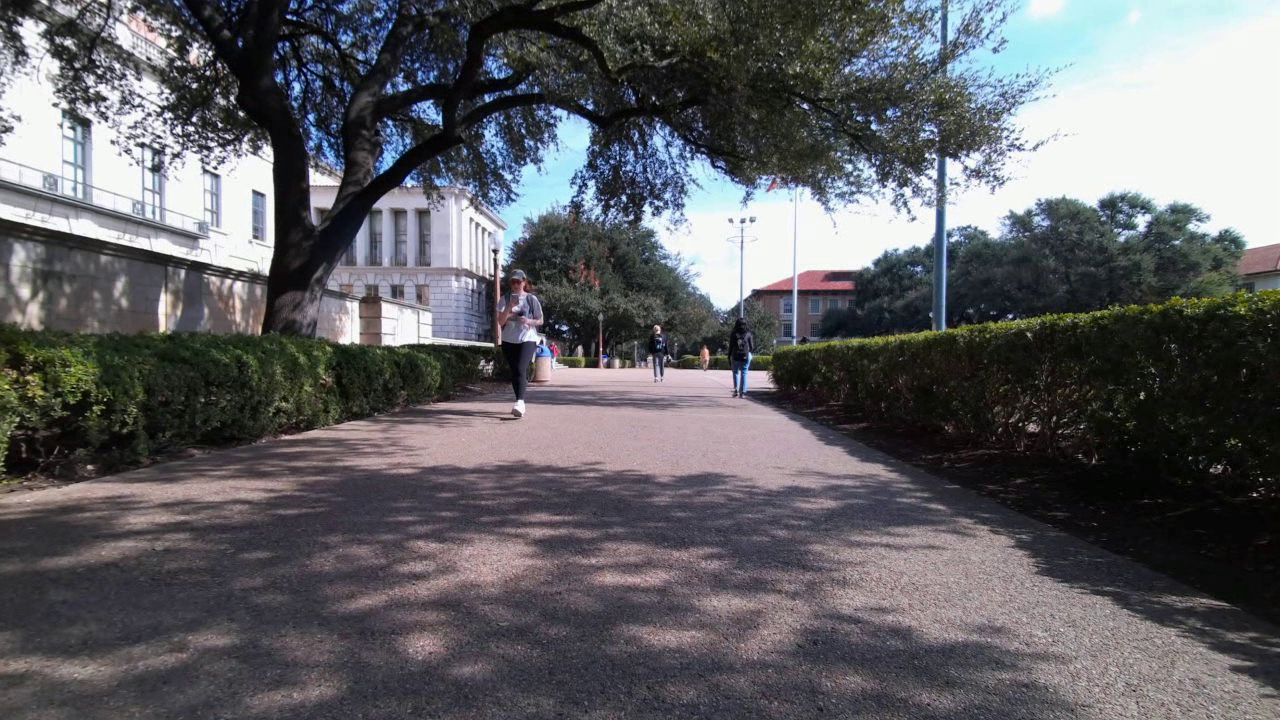}
    \end{wrapfigure}

\noindent
\user{``What do you perceive from the image?''} \\
\model{``The robot is moving forward on a sidewalk with a very low crowd density.
There is one human, a person wearing a black hat,
\underline{moving from the northwest to the southwest at a slow speed.}''} \\
\gt{``There is a human on the left side of the sidewalk
\underline{moving from the northwest to the southwest} at a far distance,
wearing a white shirt.''} \\
\user{``What do you predict these humans will do next?''} \\
\model{``The person with a black hat
\underline{walking from the northwest to the southwest}
will \underline{continue moving} at a moderate distance.''} \\
\gt{``The human will \underline{continue walking from the northwest to the southwest.}''} \\
\user{``What should the robot do?''} \\
\model{\action{``\underline{The robot should continue moving forward at a moderate speed.}''}} \\
\gt{\action{``\underline{The robot should continue moving forward at a moderate speed.}''}}
\end{tcolorbox}
\caption{Visualizations: E-SocialNav accurately captures social-compliance cues from the image.}
\label{fig:vis}
\end{figure*}

\begin{table*}[h!]
\centering
\caption{Experimental results across different VT and SLM configurations. Best performance is bolded.}
\label{tab:VT-LLM}
\resizebox{\textwidth}{!}{
\begin{tabular}{lccccccc}
\toprule
&VT &  SLM& BERTScore-P$\uparrow$  & BERTScore-R$\uparrow$  & BERTScore-F1$\uparrow$ & SBERT-cos$\uparrow$ & SMS$\uparrow$ \\
\midrule
\multirow{5}{*}{E-SocialNav (Ours)}  & CLIP~\cite{radford2021learning} & Phi-2-2.7B~\cite{javaheripi2023phi} & 0.555    &  0.658 &0.605     &  0.803  &   0.768  \\ 
& Dino~\cite{oquab2024dinov2} & Phi-2-2.7B~\cite{javaheripi2023phi} &  0.692   & 0.663   &   0.677  &  0.801 &  0.833   \\ 
 & SigLIP~\cite{zhai2023sigmoid} & TinyLlama-1.1B-Chat-v1.0~\cite{zhang2024tinyllama} &   0.473   &  0.576    &  0.523&  0.733  & 0.789    \\  
 & SigLIP~\cite{zhai2023sigmoid} & stablelm-2-zephyr-1\_6b~\cite{bellagente2024stable} &  0.700    & 0.622     &  0.661 &  0.788  &    0.837  \\  
 & SigLIP~\cite{zhai2023sigmoid} & Phi-2-2.7B~\cite{javaheripi2023phi} & \textbf{0.706}   &  \textbf{0.671}   &    \textbf{0.688} &   \textbf{0.814}  &  \textbf{0.846}   \\ 
\bottomrule
\end{tabular}
}
\end{table*}

\begin{figure}[htbp!]
\centering
\includegraphics[width=0.5\textwidth]{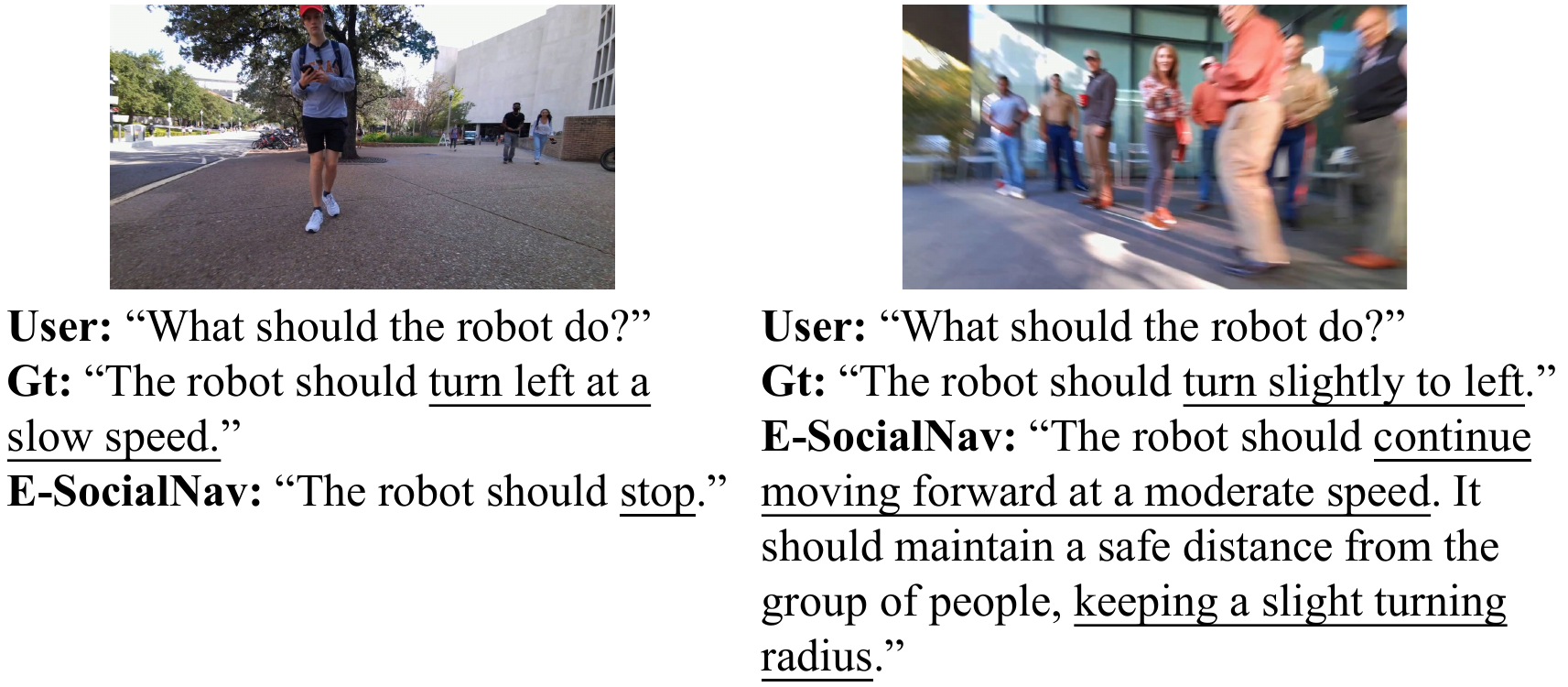}
\caption{Visualization of failure cases. Gt: Ground truth.}
\label{fig:failure}
\end{figure}

Socially compliant navigation aims to generate trajectories that are not only efficient and collision-free but also consistent with human social norms. Conceptually, this can be viewed as optimizing a composite objective that balances three factors: (i) progress toward the goal, (ii) safety in avoiding collisions and maintaining appropriate distances from obstacles, and (iii) adherence to socially compliant behaviors. The overall framework of E-SocialNav is illustrated in Figure~\ref{fig:model}. E-SocialNav consists of two training phases:

\noindent\textbf{Supervised Fine-tuning (SFT):} We optimize only the projector. To enhance robust multimodal understanding, we employ multi-dialog datasets in which each training sample contains multi-turn conversations paired with corresponding images. This design enables the model to learn not only accurate perception but also context-aware reasoning across dialogue turns.

Formally, given an image $I$ and $T$ dialogue turns $\{(x_t,y_t)\}_{t=1}^{T}$,
we encode $I$ with a vision tower (VT) and a projector to obtain visual tokens $v(I)$,
and form the multimodal context $x_t = [c_t;\, v(I)]$ by concatenating textual context $c_t$ and $v(I)$.
Let $y_{t,1:N_t}$ be the tokenized assistant response and $y_{t,<n}=(y_{t,1},\ldots,y_{t,n-1})$.

The SFT objective averages the next-token negative log-likelihood over response tokens:
\[
\mathcal{L}_{\mathrm{SFT}}(\theta)
= \frac{1}{\sum_{t=1}^{T} N_t}
  \sum_{t=1}^{T}\sum_{n=1}^{N_t}
  \big[-\log \pi_\theta\!\big(y_{t,n}\mid x_t,\, y_{t,<n}\big)\big],
\]
where $\pi_\theta$ denotes the conditional probability distribution defined by the model parameters $\theta$.
The loss is computed only on assistant responses; prompts and image tokens are excluded.

\noindent\textbf{Direct Preference Optimization (DPO):} For each input, two candidate responses are provided. The \textit{chosen response} is the human-annotated ground-truth answer, considered the most reliable. The \textit{rejected response} is constructed by modifying the ground-truth answer with localized errors. Some examples are given in Figure~\ref{fig:dpo-visual}:


Formally, for each input context $x_t$ (including visual tokens from $I$), 
we pair a \emph{chosen} response $y_t^{+}$ and a \emph{rejected} response $y_t^{-}$.
The sequence log-likelihoods are computed by summing token log-probabilities 
over supervised positions:
\begin{align}
\ell_\theta^{+}(t) &= \sum_{n} \log \pi_\theta\!\big(y_{t,n}^{+}\mid x_t,\, y_{t,<n}^{+}\big), \\
\ell_\theta^{-}(t) &= \sum_{n} \log \pi_\theta\!\big(y_{t,n}^{-}\mid x_t,\, y_{t,<n}^{-}\big).
\end{align}
Let us define the log-likelihood advantage
\[
\Delta_\theta(t) = \ell_\theta^{+}(t) - \ell_\theta^{-}(t).
\]

The DPO objective is the average binary logistic loss:
\begin{equation}
\mathcal{L}_{\text{DPO}}(\theta)
= - \frac{1}{T} \sum_{t=1}^{T} \log \sigma\!\big(\beta\,\Delta_\theta(t)\big),
\end{equation}
where $\sigma(\cdot)$ is the logistic sigmoid and $\beta>0$ is an 
inverse-temperature hyperparameter controlling the sharpness of preference learning.
In practice, we set $\beta=0.1$, which provides stable gradients without 
over-amplifying preference margins.

\section{Experiments}
\subsection{Experimental Settings}

The projector is a two-layer MLP. For evaluation, we use BERTScore, SBERT-cosine, and Sentence Mover’s Similarity (SMS), as they emphasize semantic similarity. In addition, we evaluate action accuracy (AA) to assess decision-level performance, defined as the proportion of samples in which the predicted action exactly matches the ground truth action.

Based on the SNEI dataset~\cite{payandeh2024social}, which is derived from SCAND~\cite{karnan2022socially} and MuSoHu~\cite{nguyen2023toward}, we construct a multi-dialog dataset comprising 325 egocentric video-derived samples, each paired with five-turn conversations. Among these, 60 samples are randomly selected for testing, while the remaining 265 are used for training. We also derive a DPO dataset (see Section~\ref{sec:method}, \textbf{Direct Preference Optimization (DPO)}). Training follows a two-stage schedule on four A100 GPUs and finishes in under one hour. Stage~I updates the projector for 20 epochs with a learning rate of $5\times 10^{-5}$ and a warm-up ratio of 0.03 using FlashAttention-2. Stage~II applies DPO for 5 epochs with the same settings.

\subsection{Experimental Results}
\noindent\textbf{Accuracy.} With GPT-4v deprecated, GPT-4o serves as the GPT baseline. From Table~\ref{tab:model_comparison}, both Claude and GPT-4o exhibit limited social compliance, while E-SocialNav aligns more closely with human annotations, achieving higher semantic-similarity scores and action accuracy. In the low-data regime (265 images for training), SFT performs best when the backbone is frozen and only the projector is trained (Stage~I). Adding Stage~II DPO fine-tuning further improves performance in terms of both semantic similarity to human annotations and action accuracy.

\noindent\textbf{Efficiency.} E-SocialNav builds on the compact 2.7B Phi-2 backbone, with Stage~I updates only the projector and Stage~II applies lightweight DPO. These choices keep training compute modest and reduce inference memory and latency, enabling deployment on resource-constrained hardware. 

\noindent\textbf{Visualizations.} As can be seen from Figure~\ref{fig:vis}, the proposed E-SocialNav produces responses that closely align with human annotations, reflecting both higher semantic fidelity and stronger social compliance.

\noindent\textbf{Variations on VT and SLM.}
We conduct experiments by varying both the VTs and SLMs. The VTs evaluated include CLIP~\cite{radford2021learning}, DINO~\cite{oquab2024dinov2}, and SigLIP~\cite{zhai2023sigmoid}, while the SLMs considered are Phi-2-2.7B~\cite{javaheripi2023phi}, TinyLlama-1.1B-Chat-v1.0~\cite{zhang2024tinyllama}, and StableLM-2-Zephyr-1.6B~\cite{bellagente2024stable}. Among all combinations, SigLIP paired with Phi-2-2.7B consistently achieves the best performance across all evaluation metrics (Table~\ref{tab:VT-LLM}).

\noindent\textbf{Failure Analysis and Future Works.}
As shown in Figure~\ref{fig:failure}, E-SocialNav recommends ``stop'', whereas the ground-truth annotation prescribes ``turn left at a slow speed''. This divergence reflects the inherently conservative bias often adopted during human annotation and underscores the difficulty of establishing a universally valid social standard for navigation. Moving forward, we plan to (i) conduct subjective user studies and (ii) construct a fine-grained, large-scale benchmark dataset that captures diverse cultural norms and situational contexts. Such efforts aim to provide a more balanced foundation to mitigate annotation bias and advance the development of socially compliant navigation models that are universally adaptable.

\section{Conclusions}

In this paper, we first examined the effectiveness of off-the-shelf LLMs, including Claude and GPT-4o, and found that they exhibit limited social compliance in navigation tasks. To address this limitation, we proposed E-SocialNav, a lightweight model designed for socially compliant navigation under small-data settings. By adopting a two-stage training pipeline consisting of SFT and DPO, E-SocialNav achieves higher semantic similarity to human annotations and higher action accuracy than zero-shot baselines. Notably, our E-SocialNav leverages a SLM with LoRA and projector fine-tuning for efficient adaptation, enabling faster response times, reduced energy consumption, and more practical deployment.

\section*{Acknowledgments}
This research was supported by the Japan Society for the Promotion of Science (JSPS) KAKENHI Grant Number 24K20787.

\bibliographystyle{IEEEbib}
\bibliography{strings,refs}

\end{document}